\documentclass[letterpaper, 10 pt, conference]{ieeeconf}  % Comment this line out if you need a4paper

\IEEEoverridecommandlockouts                              % This command is only needed if 
\usepackage{cite}
\usepackage{amsmath,amssymb,amsfonts}
\usepackage{algorithmic}
\usepackage{graphicx}
\usepackage{textcomp}
\usepackage{xcolor}
\def\BibTeX{{\rm B\kern-.05em{\sc i\kern-.025em b}\kern-.08em
    T\kern-.1667em\lower.7ex\hbox{E}\kern-.125emX}}
\usepackage[linesnumbered,ruled,vlined]{algorithm2e}
\usepackage{bbm}
\usepackage{bm}
\usepackage{booktabs}
\usepackage{multirow}
\usepackage{graphicx}
\usepackage{url}
\usepackage{tabularx}
\usepackage{tabu}
\usepackage{caption}
\usepackage{subcaption}
\makeatletter
\let\NAT@parse\undefined
\makeatother
\definecolor{rblue}{rgb}{0,0.5,1}
\definecolor{hollywoodcerise}{rgb}{0.96, 0.0, 0.63}
\definecolor{lasallegreen}{rgb}{0.03, 0.47, 0.19}
\definecolor{hanpurple}{rgb}{0.32, 0.09, 0.98}
\definecolor{green(pigment)}{rgb}{0.0, 0.65, 0.31}
\usepackage[pagebackref=false,breaklinks=true,colorlinks,bookmarks=false]{hyperref}
\hypersetup{colorlinks,linkcolor={red},citecolor={hanpurple},urlcolor={magenta}} 
\title{\LARGE \bf
Exploring Video-Based Driver Activity Recognition under Noisy Labels
}

\author{Linjuan Fan$^{1,+}$, Di Wen$^{1,+}$, Kunyu Peng$^{1,*}$,  Kailun Yang$^{2}$, Jiaming Zhang$^{1}$, Ruiping Liu$^{1}$, Yufan Chen$^{1}$,\\Junwei Zheng$^{1}$, Jiamin Wu$^{3}$, Xudong Han$^{4}$, and Rainer Stiefelhagen$^{1}$%
\thanks{The project is funded by the Deutsche Forschungsgemeinschaft (DFG, German Research Foundation) – SFB 1574 – 471687386. This work was supported in part by the SmartAge project sponsored by the Carl Zeiss Stiftung (P2019-01-003; 2021-2026), the MWK through the Cooperative Graduate School Accessibility through AI-based Assistive Technology (KATE) under Grant BW6-03, and in part by the BMBF through a fellowship within the IFI program of the German Academic Exchange Service (DAAD), in part by the HoreKA@KIT supercomputer partition, in part by the National Natural Science Foundation of China (Grant No. 62473139), in part by the Hunan Provincial Research and Development Project (Grant No. 2025QK3019), and in part by the Open Research Project of the State Key Laboratory of Industrial Control Technology, China (Grant No. ICT2025B20).}
\thanks{$^{1}$The authors are with the Institute for Anthropomatics and Robotics, Karlsruhe Institute of Technology, Germany (email: firstname.lastname@kit.edu).}
\thanks{$^{2}$The author is with the School of Artificial Intelligence and Robotics and the National Engineering Research Center of Robot Visual Perception and Control Technology, Hunan University, Changsha, China (email: kailun.yang@hnu.edu.cn).}
\thanks{$^{3}$The author is with Shanghai AI Lab, (email: jiaminwu@cuhk.edu.hk).}
\thanks{$^{4}$The author is with the University of Sussex, (email: xh218@sussex.ac.uk).}
\thanks{$^{*}$Corresponding author. $^{+}$Shared first author.}
}

\begin{document}

\maketitle
\thispagestyle{empty}
\pagestyle{empty}

%%%%%%%%%%%%%%%%%%%%%%%%%%%%%%%%%%%%%%%%%%%%%%%%%%%%%%%%%%%%%%%%%%%%%%%%%%%%%%%%
\begin{abstract}
As an open research topic in the field of deep learning, learning with noisy labels has attracted much attention and grown rapidly over the past ten years. Learning with label noise is crucial for driver distraction behavior recognition, as real-world video data often contains mislabeled samples, impacting model reliability and performance. However, label noise learning is barely explored in the driver activity recognition field. In this paper, we propose the first label noise learning approach for the driver activity recognition task. Based on the cluster assumption, we initially enable the model to learn clustering-friendly low-dimensional representations from given videos and assign the resultant embeddings into clusters. We subsequently perform co-refinement within each cluster to smooth the classifier outputs. Furthermore, we propose a flexible sample selection strategy that combines two selection criteria without relying on any hyperparameters to filter clean samples from the training dataset. We also incorporate a self-adaptive parameter into the sample selection process to enforce balancing across classes. A comprehensive variety of experiments on the public Drive\&Act dataset for all granularity levels demonstrates the superior performance of our method in comparison with other label-denoising methods derived from the image classification field. The source code is available at \url{https://github.com/ilonafan/DAR-noisy-labels}.

\end{abstract}

%%%%%%%%%%%%%%%%%%%%%%%%%%%%%%%%%%%%%%%%%%%%%%%%%%%%%%%%%%%%%%%%%%%%%%%%%%%%%%%%
\section{Introduction}
The advancement of the automobile industry has significantly enhanced human daily life in recent years. It has not only shortened transportation time but also improved global communication and connectivity. 
Nevertheless, a study by the World Health Organization (WHO) revealed that traffic accidents accounted for about $2.2\%$ of all deaths in 2020~\cite{world2019world}, totaling $1.35$ million fatalities.
Recognition of the distracted driver behaviors is thereby a more and more important research direction, where deep-learning-based human action recognition approaches are well employed, \textit{e.g.}, using 3D convolutional approaches~\cite{feichtenhofer2020x3d} or spatiotemporal transformer-based approaches~\cite{li2022mvitv2}.

Existing driver action recognition task usually relies on abundant data with accurate annotations~\cite{peng2022transdarc,roitberg2020open,vats2022key,doshi2022federated}. 
The presence of corrupted labels due to mislabeling, ambiguous annotations, or limitations of the labeling process is common and inevitable in real-world data. 
Research in the field of driver activity recognition under noisy labels is thereby crucial for improving road safety and reducing traffic accidents. 
Accurately identifying driver behaviors, even with imperfect labels, can help in developing advanced driver assistance systems and ensuring more reliable and effective interventions.
Learning directly from noisy labels can severely degrade the accuracy and performance of action recognition models, since deep learning models are prone to fitting the label noise, leading to poor performance during test time. 

Existing works to alleviate the noisy label disturbance are mostly from other fields, \textit{e.g.}, image classification.
Many existing research works investigate and develop efficient learning approaches to make the models more robust and adaptable to the label noise~\cite{li2024noisy, xia2021robust, liu2020earlylearning, li2021rrl}. 
Besides, recent noise-robust methods usually incorporate sample selection into the learning frameworks. Sample selection strategies rely on hyperparameters as thresholds or controls for the selection process~\cite{li2024noisy, li2021rrl, li2020dividemix}. 
These hyperparameters require rigorous tuning, increasing the complexity and computational cost. 
However, the driver activity recognition under the noisy label challenge has been nearly ignored by the community in the past.

This study investigates a label noise-robust learning method for driver activity recognition. We establish a new benchmark by applying human activity recognition backbones and noise-handling methods to the Drive\&Act dataset~\cite{drive_and_act_2019_iccv}.

We propose a novel approach that learns spatiotemporal features via contrastive learning and produces clustering-friendly low-dimensional representations using a deep clustering module. After unsupervised learning, embeddings are clustered, and the SoftMax classifier outputs are co-refined.

Our method introduces a hyperparameter-free sample selection strategy, combining Jensen-Shannon divergence-based refinement and cluster-specific thresholds. A self-adaptive parameter enforces class balance during selection. Soft pseudo-labels are generated for noisy samples, enabling semi-supervised learning on clean and pseudo-labeled subsets.
Our approach achieves state-of-the-art performance on the new benchmark for driver activity recognition under label noise scenarios.

\section{Related Work}
\subsection{Noisy Label Learning}
Over the past decade, significant research has focused on learning with noisy labels, particularly in image classification~\cite{zhang2024badlabel,sheng2024adaptive,chen2024label}. Some methods estimate the noise transition matrix and design architectures to model label noise~\cite{lee2019robust, hendrycks2019using}. Others mitigate noise memorization by incorporating regularization into objective functions~\cite{zhang2018mixup,xia2021robust,liu2020earlylearning}.

Recent approaches emphasize sample selection strategies to filter clean samples, training only on those with smaller losses~\cite{jiang2018mentornet, malach2017decoupling,han2018co}. While effective, these methods often underutilize the dataset. To address this, advanced methods integrate semi-supervised learning, treating filtered samples as labeled and generating pseudo-labels for unfiltered samples~\cite{li2021rrl, li2020dividemix, karim2022unicon}, achieving superior performance~\cite{li2024noisy}.
\subsection{Driver Activity Recognition}
\label{related_works:drive_act_dataset}
Deep learning has shown promising performances on various of areas, \textit{e.g.}, natural language processing~\cite{chang2024survey,hu2020improving}, video understanding~\cite{peng2024referring,liu2022video}, and human activity recognition~\cite{peng2024navigating,fu2023component,lerch2024unsupervised}. Driver activity recognition is a subdirection under human activity recognition, which also moves from the approaches relying on handcrafted features into end-to-end learning pipelines~\cite{roitberg2022comparative,hasan2024vision,yuan2023peer}.
Traditional driver behavior recognition relies on handcrafted features with classifiers like SVMs~\cite{ohn2014head_eye_hand} or random forests~\cite{xu2014realtime_random_forests}.
With the rise of CNNs~\cite{he2016resnet}, deep learning pipelines dominate driver activity analysis, employing top CNNs~\cite{wang2018nonlocal}, spatiotemporal models like I3D~\cite{carreira2017quo}.
Recent works explore advanced techniques to achieve driver activity recognition. Zhao~\textit{et al.}\cite{zhao2021driver} adopt spatial attention mechanisms, while Peng~\textit{et al.}~\cite{peng2022transdarc} illustrate the strength of transformers when dealing with the driver activity recognition task.
In this work, we build the first benchmark to achieve noisy label learning in the driver activity recognition field. Unlike conventional noisy label learning problems that primarily deal with classification noise in static images, our approach is tailored to handle noisy annotations in video-based tasks, incorporating spatiotemporal contrastive learning and clustering-based refinement to enhance feature learning under label noise. Additionally, we incorporate a class-balancing strategy to mitigate the long-tailed distribution of driver activities, a critical aspect of this domain. These design choices explicitly address the complexities of driver behavior recognition.

\section{Methodology}
\begin{figure*}[!t]
    \centering 
    \includegraphics[width=\textwidth]{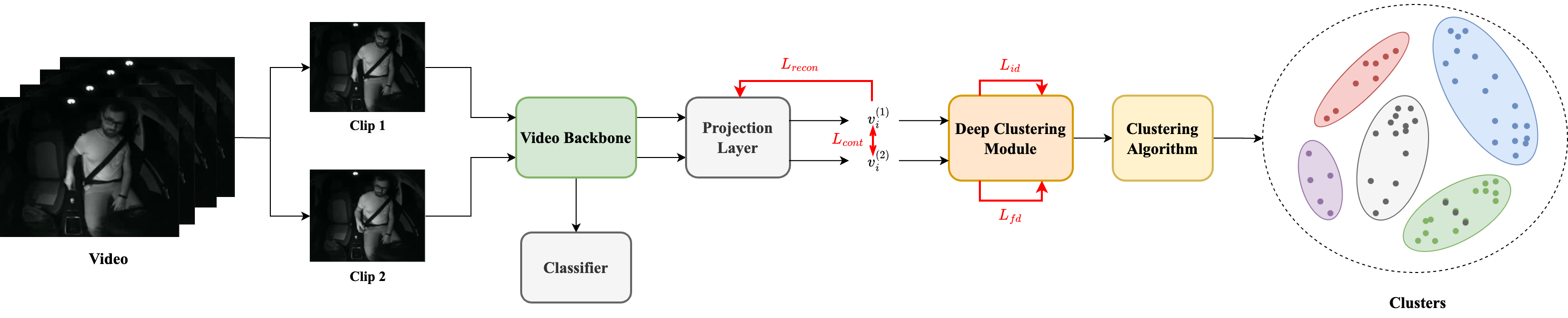}
    \caption{Proposed framework for unsupervised clustering-friendly representation learning: Two augmented clips from an input video are passed through a video backbone and a linear projection layer to generate normalized low-dimensional embeddings. A deep clustering module, combining instance discrimination ($L_{id}$) and feature decorrelation ($L_{fd}$), ensures clustering-friendly representations. The embeddings are then assigned to clusters using a clustering algorithm.
    }

    \label{fig:architecture}
    \end{figure*}

In this section, we introduce our robust learning framework for label denoising, consisting of two main components. First, we describe the unsupervised learning of clustering-friendly representations (Figure~\ref{fig:architecture}). Next, we outline the sample selection process and the semi-supervised learning procedure. The full algorithm is provided in pseudo-code in Algorithm~\ref{algo:proposed_method}. 

\subsection{Unsupervised Clustering-Friendly Representation Learning}
We aim at using contrastive learning to enhance spatiotemporal representations by encouraging high similarity within the same video while distinguishing between different videos, resulting in robust and discriminative embeddings. Instance discrimination builds upon this by treating each instance as a distinct class, promoting finer-grained distinctions that lead to clustering-friendly representations. Applying instance discrimination after contrastive learning further improves clustering performance. Once these discriminative and unsupervised features are obtained, clean sample selection is used to identify reliable data points, which then serve as priors for generating pseudo-labels for samples suspected of containing label noise.
\label{method:UCRL}

    \subsubsection{Spatiotemporal Contrastive Learning}
    \label{method:spatiotemporal_learning}
        \par
        With a noisy dataset $D=\{(\mathbf{x}_i, \mathbf{y}_i)\}_{i=1}^n$, where $\mathbf{x}_i$ denotes the $i$-th input video and $\mathbf{y}_i \in\{1, \ldots, K\}$ denotes its observed label, we aim to learn both spatially and temporally persistent features from input videos by encouraging the similarity of feature representations across different temporal clips of the same video. We leverage the unsupervised contrastive learning approach~\cite{feichtenhofer2021largescale} that extends the idea of SimCLR~\cite{chen2020simple} to the space-time domain.

        Given a mini-batch of $N_b$ videos, we randomly sample two clips out of a fixed number of frames of the raw video $\mathbf{x}_i$ and apply a random combination of spatial augmentations to produce a pair of augmented clips. The clips are encoded as high-dimensional embeddings by a video backbone to obtain spatiotemporal representations. These high-dimensional embeddings are then projected into a low-dimensional subspace by a linear projection layer $\mathbf{W}_{\mathrm{v}}$, resulting in a pair of normalized low-dimensional embeddings $\mathbf{v}_i$ and $\mathbf{v}_{j(i)}$.
        
        A contrastive loss is used to maximize the similarity between the pair of embeddings from the same input video $\mathbf{v}_i$ and $\mathbf{v}_{j(i)}$ while minimizing the similarity of the embeddings from different videos within the mini-batch. The objective function is defined as:
            \begin{equation}
            \label{eq:contrastive_loss}
                \mathcal{L}_{cont}=-\sum_{i=1}^{N_b} log \frac{exp (\mathbf{v}_i \cdot \mathbf{v}_{j(i)} / \tau_1)}{\sum_{k=1}^{2 {N_b}} \mathrm{1}_{i \neq k} exp (\mathbf{v}_i \cdot \mathbf{v}_k / \tau_1)},
            \end{equation}
        where the similarity is calculated by the inner product between the embeddings, ${N_b}$ denotes the size of a mini-batch, and $\tau_1$ is a scaling temperature hyperparameter.

        Additionally, we utilize a linear decoder $\mathbf{W}_{\mathrm{d}}$ to reconstruct the high-dimensional embedding $\mathbf{u}_i = f_\theta(\mathbf{x}_i)$ from the low-dimensional representation $\mathbf{v}_i$, where $f_\theta$ denotes the video encoder parameterized by $\theta$. The decoder aims to preserve the mutual information between the latent spaces. The L2 reconstruction loss is defined as:

            \begin{equation}
            \label{eq:recon_loss}
                \mathcal{L}_{recon}=\sum_{i=1}^{2N_b}\|\mathbf{u}_i - \mathbf{W}_{\mathrm{d}} \mathbf{v}_i\|^2.
            \end{equation}

    \subsubsection{Label Noise Agnostic Clustering Module}
    \label{method:deep_clustering_module}
        \par
        One of the core principles underlying our proposed method is the cluster assumption, which posits that points within the same cluster are likely to belong to the same class~\cite{chapelle2006semi-supervised}. To explicitly enhance the performance of the subsequent clustering task, we integrate a deep clustering module into the representation learning framework. This module is employed to learn representations that are more conducive to effective clustering.

        \noindent \textbf{Instance discrimination}. 
        The deep clustering module first applies the instance discrimination method ~\cite{wu2018unsupervised} to capture the similarities among instances. In this method, we assume that each instance has a unique class of its own. The primary objective is to distinguish each individual instance from other instances in the dataset. We leverage the normalized low-dimensional embedding $\mathbf{v}$ as the weight vector for each class. Given a dataset of $n$ input samples, the probability of $i$-th sample being assigned to $k$-th class is defined as:
            \begin{equation}
                P(k|\mathbf{v}_i)=\frac{\exp (\mathbf{v}_{k} \cdot \mathbf{v}_i / \tau_2)}{\sum_{j=1}^{n} \exp (\mathbf{v}_{j} \cdot \mathbf{v}_i / \tau_2)},
            \end{equation}
        where $\mathbf{v}_{k} \cdot \mathbf{v}_i$ measures how well the embeddings of $\mathbf{x}_i$ matches $k$-th class. $\tau_2$ is a temperature hyperparameter to control the concentration level of the distribution~\cite{wu2018unsupervised}. 

        A negative log-likelihood loss is employed to equivalently maximize the joint probability $\prod_{i=1}^n P(i \mid \mathbf{v}_i)$ over the entire dataset and is defined as:
            \begin{equation}
                \mathcal{L}_{id}=-\sum_{i=1}^n \log P(i|\mathbf{v}_i).
            \end{equation}
        
        \noindent \textbf{Feature decorrelation}.
        The deep clustering module incorporates the feature decorrelation method to minimize correlations within features, enabling the learning of independent features with reduced redundant information~\cite{tao2021clusteringfriendly}.

        We transpose the low-dimensional embeddings to obtain a set of feature vectors $\{\mathbf{f}_l\}_{l=1}^d$, where $d$ equals the dimensionality of the embeddings. We then normalize the feature vectors by the L2 norm to transform them into unit vectors. 
                
        Analogous to the instance discrimination method, a softmax-formulated constraint is applied to encourage the correlation between a feature vector and itself to converge to $1$, while the correlation between a feature vector and other vectors converges to ${-}1$ or $0$~\cite{tao2021clusteringfriendly}. The constraint is defined as:
            \begin{equation}
                Q(\mathbf{f}_l)=\frac{\exp (\mathbf{f}_{l} \cdot \mathbf{f}_l / \tau_3)}{\sum_{m=1}^{d} \exp (\mathbf{f}_{m} \cdot \mathbf{f}_l / \tau_3)},
            \end{equation}
        where $\mathbf{f}_{m} \cdot \mathbf{f}_l$ measures how $i$-th feature vector correlates to another feature vector. $\tau_3$ is a temperature hyperparameter. 

        A negative log-likelihood is adopted to equivalently maximize the decorrelation constraints over all the features and is defined as:      
            \begin{equation}
                \mathcal{L}_{fd}=-\sum_{l=1}^d \log Q(\mathbf{f}_l).
            \end{equation}

\begin{algorithm}[t]
    \caption{Pseudo-code of our proposed method}
    \label{algo:proposed_method}
    \DontPrintSemicolon
    \SetNoFillComment
    \SetKwFunction{SGD}{SGD}
    \SetKwFunction{Beta}{Beta}
    \SetKwFunction{Clustering}{Clustering-Algorithm}
    \SetKwFunction{Softmax}{softmax}
    \SetKwFunction{JSD}{JSD}
  
    \KwIn{noisy dataset $D=\{(\mathbf{x}_i, y_i)\}_{i=1}^n$, model parameters $\theta$}
    \KwOut{robust model parameters $\theta$}
        \For{$t = 0$ \KwTo $T_1 - 1$}{ 
            Sample a mini-batch of $N_b$ pairs of augmented clips $\{\mathbf{x}_i\}_{i=1}^{2N_b}$ without labels from $D$\;
            
            $\mathbf{v}_i=\frac{f_\theta(\mathbf{x}_i)}{\|f_\theta(\mathbf{x}_i)\|}$ 

            ${\mathcal{L}}=\sum_{i=1}^{2N_b} (\mathcal{L}_{cont}(\mathbf{v}_i)+\mathcal{L}_{recon}(\mathbf{v}_i)+\mathcal{L}_{id}(\mathbf{v}_i)+ \mathcal{L}_{fd}(\mathbf{v}_i))$ \;
            $\theta=$\SGD$(\mathcal{L},\theta)$ 
        }

        \For{$t = T_1$ \KwTo $T_2 -1 $}{ 
            Sample a mini-batch of $N_b$ pairs of augmented clips $\{(\mathbf{x}_i, \mathbf{y}_i)\}_{i=1}^{2N_b}$\;
            
            $\lambda \sim$ \Beta$(\alpha, \alpha)$ %
            $\mathbf{x}_i^m = \lambda \mathbf{x}_i+(1-\lambda) \mathbf{x}_{m(i)}$ \;
            
            $\mathbf{y}_i^m = \lambda \mathbf{y}_i+(1-\lambda) \mathbf{y}_{m(i)}$ %
            
            ${\mathcal{L}}=\sum_{i=1}^{2N_b} (\mathcal{L}_{cont}(\mathbf{v}_i)+\mathcal{L}_{recon}(\mathbf{v}_i)+\mathcal{L}_{ce\_mix}(\mathbf{x}_i^m, \mathbf{y}_i^m))$ \; 
            $\theta=$\SGD$(\mathcal{L},\theta)$ %
        }

        \For{$t = T_2$ \KwTo $T_{max} - 1 $}{ 
            $\{\mathbf{v}_i\}_{i=1}^{n}=\{\frac{f_\theta(\mathbf{x}_i)}{\|f_\theta(\mathbf{x}_i)\|}\}_{i=1}^{n}$ 

            $\{\mathbf{c}_i\}_{i=1}^K=$\Clustering$(\{\mathbf{v}_i\}_{i=1}^{n})$ 

            $\{\mathbf{p}_i^t\}_{i=1}^n=$\Softmax$(\{f_\theta(\mathbf{x}_i)\}_{i=1}^n)$ %
            
            $\hat{\mathbf{p}}_i^t=(1 -\beta) \mathbf{p}_i^t+ \beta \sum_{j=1}^{|C_i|} w_{i j}^t \hat{\mathbf{q}}_j^{t-1}$ 

            $\{d_i\}_{i=1}^{n}=$\JSD$(\{(\hat{\mathbf{p}}_i, \mathbf{y}_i)\}_{i=1}^{n})$ 

            $d_i \sim \sum_{j=1}^2 \phi_j N(\mu_j, \sigma_j^2)$ 

            Construct a clean subset $X^t$ based on two selection criteria (Eq. \ref{eq:clean_subset}) \;
            
            Generate soft-pseudo labels $\tilde{\mathbf{q}}^t$ for noisy samples (Eq. \ref{eq:soft_pseudo_label}) and construct a pseudo-labeled subset $U^t$ \;
            
            Repeat line 2-3 and calculate $\mathcal{L}_{cont}, \mathcal{L}_{recon}$ \;
           
            ${\mathcal{L}}=\mathcal{L}_{cont}+\mathcal{L}_{recon}+\mathcal{L}_{X}+\gamma_u \mathcal{L}_{U}$\; 

            $\theta=$\SGD$(\mathcal{L},\theta)$ %
        }
    
\end{algorithm}

\subsection{Sample Selection and Semi-Supervised Learning}
\label{method:SSSL}

    \subsubsection{Warm-up Stage}
    \label{method:warmup_stage}
    
        Before partitioning $D$, we first warm up the whole model on the training dataset for several epochs. To alleviate the memorization of noisy labels, we apply Mixup augmentation~\cite{zhang2018mixup} to mini-batches, generating virtual samples $\{(\mathbf{x}_i^m, \mathbf{y}_i^m)\}$ from the training dataset for the calculation of classification loss. 
        
        A virtual sample is constructed by linear interpolation between two randomly selected samples (indexed by $i$ and $m(i)$) from a mini-batch of $2N_b$ samples:
    
            \begin{equation}
                \begin{aligned}
                    \mathbf{x}_i^m &= \lambda \mathbf{x}_i+(1-\lambda) \mathbf{x}_{m(i)}, \\
                    \mathbf{y}_i^m &= \lambda \mathbf{y}_i+(1-\lambda) \mathbf{y}_{m(i)}, \\
                \end{aligned}
            \end{equation}
    
        where $\lambda \sim \mathrm{Beta}(\alpha, \alpha)$, with $\alpha \in (0, \infty)$. $\alpha$ is a hyperparameter that controls the strength of linear interpolation. 
    
        Given a mini-batch of $2N_b$ virtual samples $\{(\mathbf{x}_i^m, \mathbf{y}_i^m)\}_{i=1}^{2{N_b}}$, the cross-entropy classification loss is formulated as:
    
            \begin{equation}
                \mathcal{L}_{ce\_mix}=-\sum_{i=1}^{2{N_b}} \log \mathbf{p}(y_i^m ; \mathbf{x}_i^m),
            \end{equation}
        where $\mathbf{p}(y_i^m ; \mathbf{x}_i^m)$ denotes the output of the softmax classifier. The unsupervised contrastive loss (Eq.~\ref{eq:contrastive_loss}) and reconstruction loss (Eq.~\ref{eq:recon_loss}) are also integrated into the overall objective function during the warming-up stage.

    \subsubsection{Co-refinement within cluster}
    \label{method:co_refinement_within_cluster}

        We obtain the low-dimensional normalized embeddings $\{\mathbf{v}_i\}_{i=1}^{n}$ by feeding all the input videos $\{\mathbf{x}_i\}_{i=1}^{n}$ to the warmed-up model. Then we employ a clustering algorithm to assign the embeddings into clusters.
        Assuming that samples assigned to the same cluster are prone to belong to the same class, we aggregate the predictive information within each cluster to refine the softmax prediction scores of the samples assigned to it. 

        The refined prediction score $\hat{\mathbf{p}}_i^t$ of the current epoch $t$ for the $i$-th input is defined as a linear combination of the softmax classifier output $\mathbf{p}_i^t$ of current epoch and a weighted sum of the soft labels ${\mathbf{q}}_j^t$ of other samples in the assigned cluster $C_i$ from the previous epoch:
            \begin{equation}
            \label{eq:co-refinement}
                \begin{aligned}
                    \hat{\mathbf{p}}_i^t=(1 -\beta) \mathbf{p}_i^t+ \beta \sum_{j=1}^{|C_i|} w_{i j}^t \hat{\mathbf{q}}_j^{t-1}, 
                \end{aligned}
            \end{equation}
        where the weight $w_{i j}^t$ is defined as the softmax value of the negative distance between the embeddings. $\beta$ is a hyperparameter to adjust the level of co-refinement.
        
       For a clean sample, the soft label is the same as its one-hot observed label, while for a noisy sample, it matches its generated soft pseudo-label (defined in Eq.~\ref{eq:soft_pseudo_label}). We initialize the soft label using the softmax prediction score from the first epoch.

    \subsubsection{Clean Samples Selection}
    \label{method:clean_samples_selection}
        To partition the training dataset $D$, we apply two different selection criteria to jointly filter the clean samples. 
        
        \noindent \textbf{Balanced prediction score-based criterion}.
        To filter samples with higher refined prediction scores into the clean subset, we utilize the mean refined prediction score of the samples within each cluster as the threshold for that cluster. 

        Additionally, we introduce a parameter, class weight $\hat{\omega}$, to promote class balancing during the sample selection process.
        The class weight at the $t$-th epoch is defined as:
            \begin{equation}
            \label{eq:class_weight_t}
                \begin{aligned}
                    \omega_c^{t} &= -\frac{|X^{t-1}(c)|}{|D(c)|},\\
                    \hat{\omega}_c^{t} &= \frac{\omega^{t} - \min(\omega^{t})}{\max(\omega^{t}) - \min(\omega^{t})} + \epsilon,\\
                 \end{aligned}
            \end{equation}
        where $X^{t-1}(c)$ indicates the clean subset containing the filtered samples with $y_i {=} c$ from the epoch $t{-}1$. $D(c)$ denotes the set of training samples with the observed label $c$. $\epsilon$ is a small constant introduced to prevent division by zero and ensure numerical stability.

        The class weight is able to self-adapt at every epoch by recomputing the size of the constructed clean subset from the previous epoch. To initialize the class weight for the first epoch, we approximate the number of clean samples by the difference between the number of training samples and the expected number of noisy samples. 

        \noindent \textbf{JSD-based criterion}.
        We leverage Jensen-Shannon divergence (JSD) between the refined prediction score $\hat{\mathbf{p}}_i$ and the one-hot encoded observed label $\mathbf{y}_i$ for each sample to measure the dissimilarity between these two distributions, defined as:
            \begin{equation}
                \begin{aligned}
                    d_i&=\mathrm{JSD}(\hat{\mathbf{p}}_i, \mathbf{y}_i)=\frac{1}{2} \mathrm{D}(\hat{\mathbf{p}_i} \| \mathbf{m}_i)+\frac{1}{2} \mathrm{D}(\mathbf{y}_i \| \mathbf{m}_i),
                \end{aligned}
            \end{equation}
        where $\mathbf{m}_i{=}\frac{1}{2}(\hat{\mathbf{p}_i}+\mathbf{y}_i)$ denotes a mixture distribution of $\mathbf{p}_i$ and $\mathbf{y}_i$. $\mathrm{D}(.)$ represents the Kullback-Leibler divergence.

        To separate the clean samples from the noisy dataset, we assume that the JSDs $\{d_i\}_{i=1}^{n}$ of all the training samples follow a two-component Gaussian mixture model ~\cite{PERMUTER2006695}: 

            \begin{equation}
            \label{eq:gmm}
                d \sim \sum_{j=1}^2 \phi_j N(\mu_j, \sigma_j^2), 
            \end{equation}
        where $\phi_j$ represents the weight of $j$-th component and $\sum_{j=1}^2 \phi_j {=} 1$. We assume ${\mu}_1{<}{\mu}_2$.

        By jointly applying these two sample selection criteria, we filter the clean samples to construct a subset $X^t$ at the $t$-th epoch as follows:
            \begin{equation}
            \label{eq:clean_subset}
                \begin{aligned}
                    X^t=\{(\mathbf{x}_i, \mathbf{y}_i) &\mid \hat{\omega}_{y_i}^t \hat{p}_{i,y_i}^t \ge \mu_{k}\sum_{c=1}^C\hat{\omega}_c^t\} \cup \\
                    \{(\mathbf{x}_j, \mathbf{y}_j) &\mid \mathrm{P}(d_j \mid \mu_1, \sigma_1)>\mathrm{P}(d_j \mid \mu_2, \sigma_2)\}.
                \end{aligned}
            \end{equation}

    \subsubsection{Soft pseudo-labels generation for noisy samples}
    \label{method:soft_pseudo_labels_generation_for_noisy_samples}
        \par
        For each noisy sample $(\mathbf{x}_i, \mathbf{y}_i) {\in} D \setminus X^t$, we aim to generate a soft pseudo-label based on its refined prediction score $\hat{\mathbf{p}}_i^t$. 
        
        We treat the class with the maximum refined prediction score as a guessing label $\mathbf{y}_i^*{=}\arg \max_k \hat{p}_{i,k}^t$. To tolerate the inaccuracies of the sample selection process, we also take the original observed label $\mathbf{y}_i$ into account. By removing the scores of other classes, the soft pseudo-label of the $i$-th noisy sample at the $t$ epoch is formulated as:
            \begin{equation}
            \label{eq:soft_pseudo_label}
                \begin{aligned}
                    \tilde{q}_i^t= \begin{cases}\hat{p}_{i,k}^t & \text { if } k \in \{\mathbf{y}_i, \mathbf{y}_i^*\} \\ 
                    0 & \text { otherwise. }\end{cases}
                 \end{aligned}
            \end{equation}

        After the generation of soft pseudo-labels, we enable to construct a subset with pseudo-labeled samples $U^t{=}\{(\mathbf{x}_i, \tilde{q}_i^t) {\mid} \mathbf{x}_i {\in} D \setminus X^t \}$.
    
    \subsubsection{Semi-supervised learning}
    \label{method:semi_supervised_learning}
        \par
        During the semi-supervised learning stage, we employ the cross-entropy loss as the classification loss for the clean subset, while adopting the L2 loss for the pseudo-labeled subset:  

            \begin{equation}
                \begin{aligned}
                    \mathcal{L}_{X}&=-\sum_{(\mathbf{x}_i, \mathbf{y}_i)\in X} \log \mathbf{p}(\mathbf{y}_i; \mathbf{x}_i), \\
                    \mathcal{L}_{U}&=\sum_{(\mathbf{x}_i, \tilde{\mathbf{q}}_i)\in U}\|\tilde{\mathbf{q}}_i-\mathbf{p}_i\|_2^2.
                \end{aligned}
            \end{equation}

        Additionally, we integrate the unsupervised contrastive loss and reconstruction loss (Eq.~\ref{eq:contrastive_loss} \& \ref{eq:recon_loss}) into the objective function.
        To regulate the impact of the pseudo-labeled classification loss, we introduce a coefficient $\gamma_u$ that linearly increases with the number of epochs.

\section{Experiments}

\begin{table}[!tb]
\centering
    \resizebox{\columnwidth}{!}{%
    \begin{tabular}{@{}c|cccc|cccc@{}}
    \toprule
    \multirow{3}{*}{Method} & \multicolumn{4}{c|}{50\% Noise}                                                        & \multicolumn{4}{c}{70\% Noise}                                                         \\ \cmidrule(l){2-9} 
                            & \multicolumn{2}{c|}{MViTv2 \cite{li2022mvitv2}}                          & \multicolumn{2}{c|}{X3D \cite{feichtenhofer2020x3d}}        & \multicolumn{2}{c|}{MViTv2}                          & \multicolumn{2}{c}{X3D}         \\ \cmidrule(l){2-9} 
                            & Val            & \multicolumn{1}{c|}{Test}           & Val            & Test           & Val            & \multicolumn{1}{c|}{Test}           & Val            & Test           \\ \midrule
    Baseline                & 78.41          & \multicolumn{1}{c|}{82.08}          & 51.61          & 53.38          & 66.17          & \multicolumn{1}{c|}{65.89}          & 46.99          & 47.69          \\
    CDR \cite{xia2021robust}    & 77.66          & \multicolumn{1}{c|}{82.47}          & 51.13          & 53.67          & 66.48          & \multicolumn{1}{c|}{67.44}          & 40.68          & 40.47          \\
    ELR \cite{liu2020earlylearning}  & 79.88      & \multicolumn{1}{c|}{\underline{83.05}}   & 51.83          & \underline{54.15}    & 66.69      & \multicolumn{1}{c|}{68.97}          & 45.91          & 46.08          \\
    RRL \cite{li2021rrl}    & \underline{80.06}    & \multicolumn{1}{c|}{82.86}       & \underline{55.58}    & 53.91          & \underline{68.68}    & \multicolumn{1}{c|}{\underline{70.70}}    & \underline{47.05}       & \underline{48.47}              \\ \midrule
    Proposed                & \textbf{81.18} & \multicolumn{1}{c|}{\textbf{84.39}} & \textbf{57.40} & \textbf{57.62} & \textbf{70.37} & \multicolumn{1}{c|}{\textbf{71.08}} & \textbf{52.05} & \textbf{53.51} \\ \bottomrule
    \end{tabular}%
    }
    \caption{Experiments on Drive\&Act \cite{drive_and_act_2019_iccv} for fine-grained activities recognition, evaluated by top-1 accuracy.} 
    %\vskip-2ex
    %
    \label{tab:fine_grained}
\end{table}

\begin{table}[!bt]
\centering
    \resizebox{\columnwidth}{!}{%
    \begin{tabular}{@{}c|cc|cc|cc|cc|cc@{}}
    \toprule
    \multirow{2}{*}{Method} & \multicolumn{2}{c|}{Coarse Task} & \multicolumn{2}{c|}{Action}     & \multicolumn{2}{c|}{Object}     & \multicolumn{2}{c|}{Location}   & \multicolumn{2}{c}{All}         \\ \cmidrule(l){2-11} 
                            & Val             & Test           & Val            & Test           & Val            & Test           & Val            & Test           & Val            & Test           \\ \midrule
    Baseline                & 59.57           & 53.87          & 77.08          & 76.98          & 68.23          & 68.59          & 73.34          & 76.41          & 41.54              & 45.60          \\
    CDR \cite{xia2021robust}                    & 59.51           & 55.13          & 77.70          & 78.05          & 68.94          & \textbf{69.51} & 72.05          & 76.51          & 45.88              & 48.74          \\
    ELR \cite{liu2020earlylearning}                   & \underline{62.33}     & 54.03          & \underline{78.14}    & \underline{78.98}    & 69.08          & 68.08          & \underline{74.50}    & \underline{78.68}    & \underline{47.19}              & \underline{49.03}    \\
    RRL \cite{li2021rrl}                    & 60.50           & \underline{54.45}    & 77.86          & 77.01          & \textbf{69.59} & 68.81          & 73.32          & 76.65          & 43.57          & 46.48          \\ \midrule
    Proposed                & \textbf{64.41}  & \textbf{57.02} & \textbf{79.38} & \textbf{79.92} & \underline{69.52}    & \underline{69.41}    & \textbf{74.98} & \textbf{79.74} & \textbf{48.06} & \textbf{50.49} \\ \bottomrule
    \end{tabular}%
    }
    \caption{Experiment results on Drive\&Act~\cite{drive_and_act_2019_iccv} with $50\%$ label noise for coarse scenarios recognition and atomic action units triplet recognition. The video encoder is MViTv2. The evaluation metric is top-1 accuracy.}
    \label{tab:coarse_and_atomic}
\end{table}

\subsection{Evaluation Results on Drive\&Act Dataset}
\label{exp:evaluation_results}

    \noindent \textbf{Label noise simulation}.
To evaluate our method under noisy labels, we simulate label noise on the Drive\&Act~\cite{drive_and_act_2019_iccv} dataset by randomly flipping $50\%$ and $70\%$ of training sample labels to other classes with equal probability.

\noindent \textbf{Implementation details}.
We employ either a transformer-based video model MViTv2~\cite{li2022mvitv2} or a convolution-based video model X3D~\cite{feichtenhofer2020x3d} as the video backbone for representation encoding, following the official code of \texttt{PySlowFast}\footnote{\url{https://github.com/facebookresearch/SlowFast}}. 
We employ the AdamW optimizer~\cite{loshchilov2019decoupled} with an initial learning rate of $1e{-}4$ and adopt a cosine annealing learning rate scheduler. 
For our method, we build an additional linear projection layer with the output dimensionality $d{=}48$. We sample two clips out of $16$ frames with a sampling rate of $4$ from each input raw video and apply a random combination of spatial augmentations, including random $224^2$ crop, horizontal flip, color distortion, and Gaussian blur to all the sampled clips. We warm up the model for $5$ epochs and apply an agglomerative clustering algorithm~\cite{müllner2011modern} to assign the learned representations into clusters. 
For further details in hyperparameter tuning, we set $\tau_{1}{=}0.3, \tau_{2}{=}0.5, \tau_{3}{=}2.0$, $\alpha{=}8.0$. We tune $\beta{=}0.5$ for the co-refinement within a cluster and set $\epsilon{=}0.05$. 
The weight of L2 loss on the pseudo-labeled subset is initially defined as $\gamma_u{=}100$ for MViTv2, while set as $\gamma_u{=}50$ for X3D. 
All the training is conducted on an NVIDIA Quadro RTX 8000 graphic card with 50GB RAM with a batch size of $4$. All the hyperparameter selections are conducted through griding search. 
%--------------------------------------------------------
\begin{table}[!tb]
\centering
    \resizebox{0.95\columnwidth}{!}{%
    \begin{tabular}{@{}c|c|c|c|c|c@{}}
    \toprule
    Method           & Precision      & Recall         &Correction      & Val            & Test           \\ \midrule
    w/o UCRL          & 87.52              & 95.92              & 75.49              & 79.43          & 82.84          \\
    w/o class weight & 87.33          & 94.76          & 71.15          & 79.50          & 82.04          \\ \midrule
    Proposed         & \textbf{92.05} & \textbf{96.05} & \textbf{75.56} & \textbf{81.18} & \textbf{84.39} \\ \bottomrule
    \end{tabular}%
    }
    \caption{Ablation study regarding two primary components. 
    The evaluation metrics are sample selection precision, recall, noisy labels correction accuracy, top-1 validation accuracy, and top-1 test accuracy.}
    \label{tab:ablation_study}
\end{table}

\begin{figure*}[!t]
\centering
\begin{subfigure}[b]{0.2\textwidth}
    \centering
    \includegraphics[width=\textwidth]{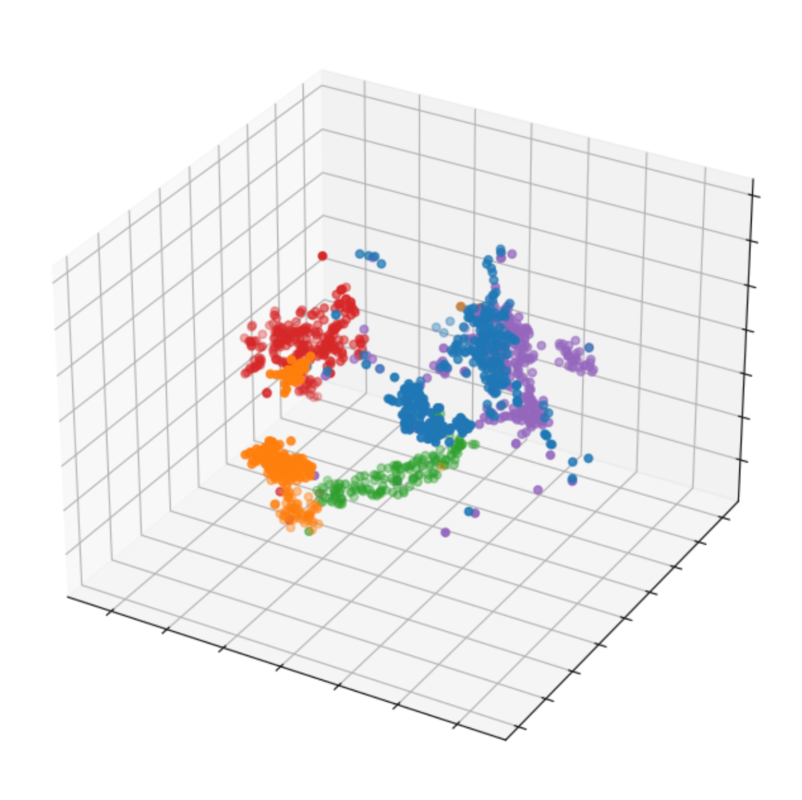}
    \caption{Head w/o UCRL}
    \label{fig:exp:wo_ucrl_head}
\end{subfigure}
\hfill
\begin{subfigure}[b]{0.2\textwidth}
    \centering
    \includegraphics[width=\textwidth]{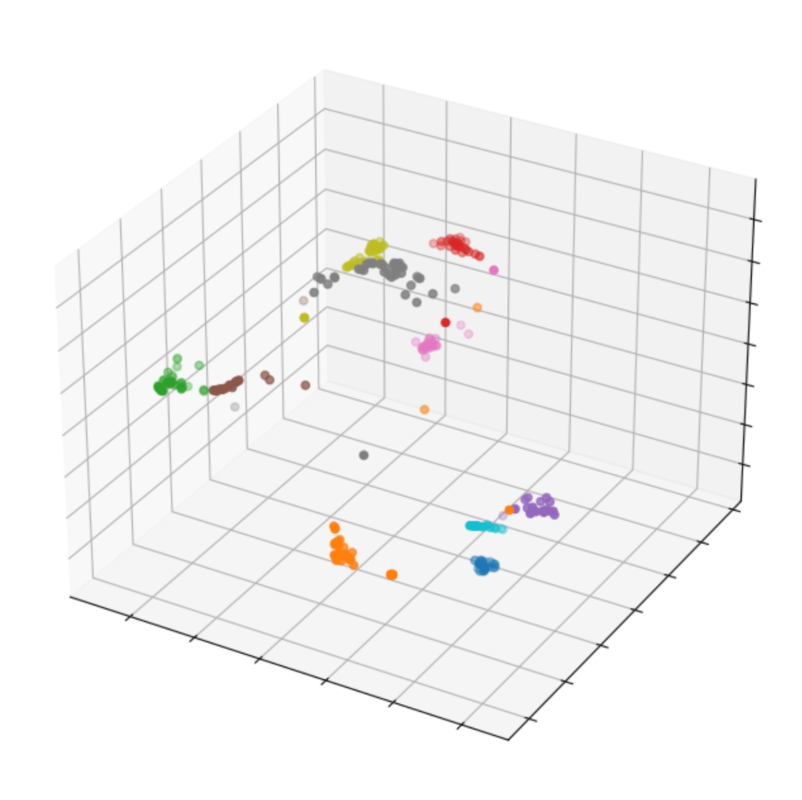}
    \caption{Tail w/o UCRL}
    \label{fig:exp:wo_ucrl_tail}
\end{subfigure}
\hfill
\begin{subfigure}[b]{0.2\textwidth}
    \centering
    \includegraphics[width=\textwidth]{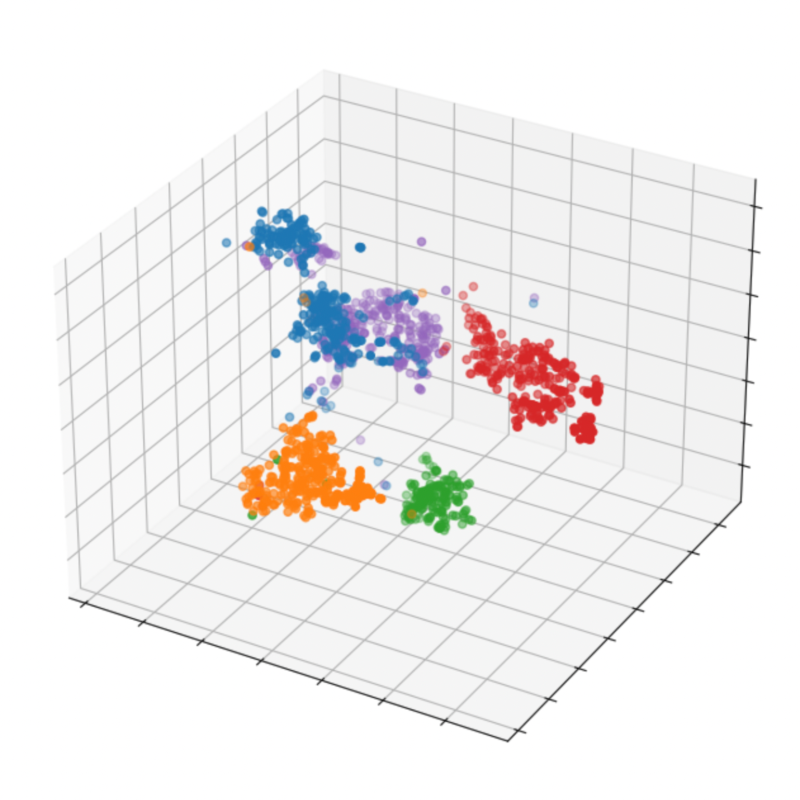}
    \caption{Head w/ UCRL}
    \label{fig:exp:w_ucrl_head}
\end{subfigure}
\hfill
\begin{subfigure}[b]{0.2\textwidth}
    \centering
    \includegraphics[width=\textwidth]{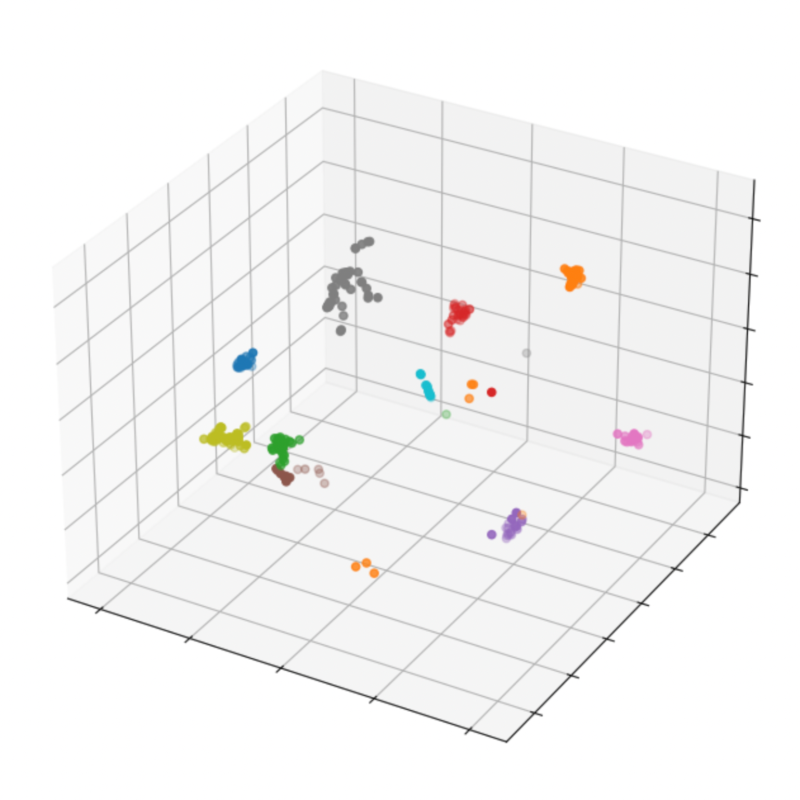}
    \caption{Tail w/ UCRL}
    \label{fig:exp:w_ucrl_tail}
\end{subfigure}
\caption{T-SNE~\cite{JMLR:v9:vandermaaten08a} visualization of learned representations of head/tail classes after the entire training process w/ and w/o UCRL.}
\label{fig:exp:ucrl_w/wo}
\end{figure*}

\begin{figure*}[!t]
\centering
\begin{subfigure}[b]{0.2\textwidth}
    \centering
    \includegraphics[width=\textwidth]{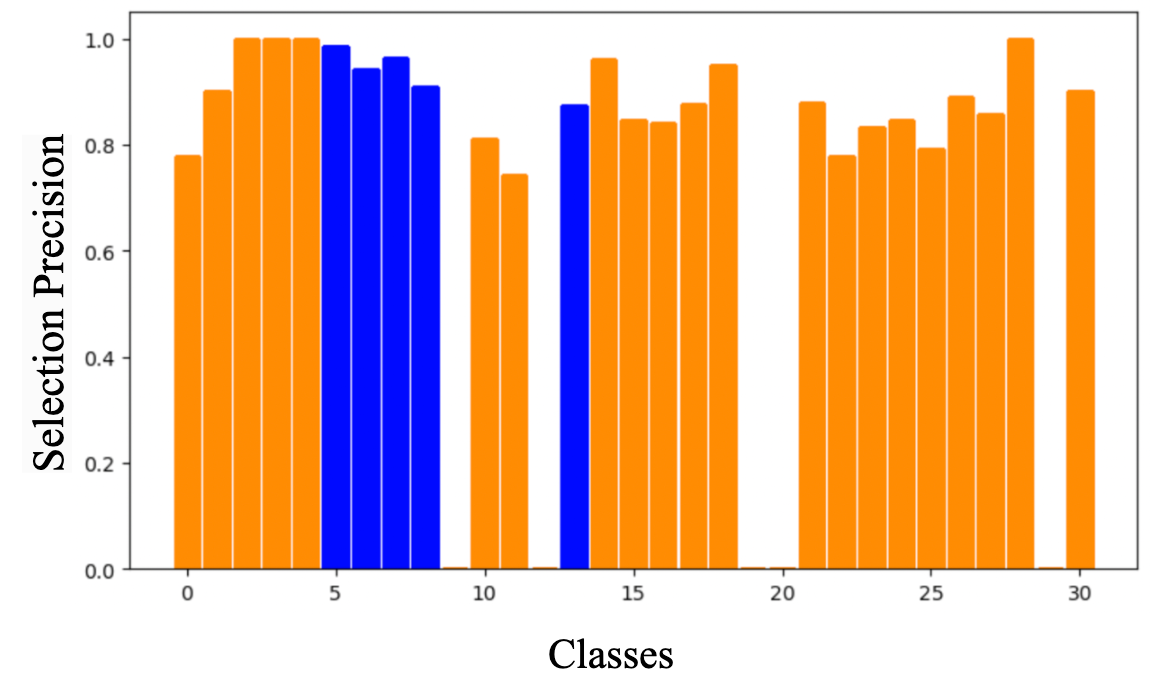}
    \caption{Prec. w/o cls weight}
    \label{fig:exp:precision_wo_cw}
\end{subfigure}
\hfill
\begin{subfigure}[b]{0.2\textwidth}
    \centering
    \includegraphics[width=\textwidth]{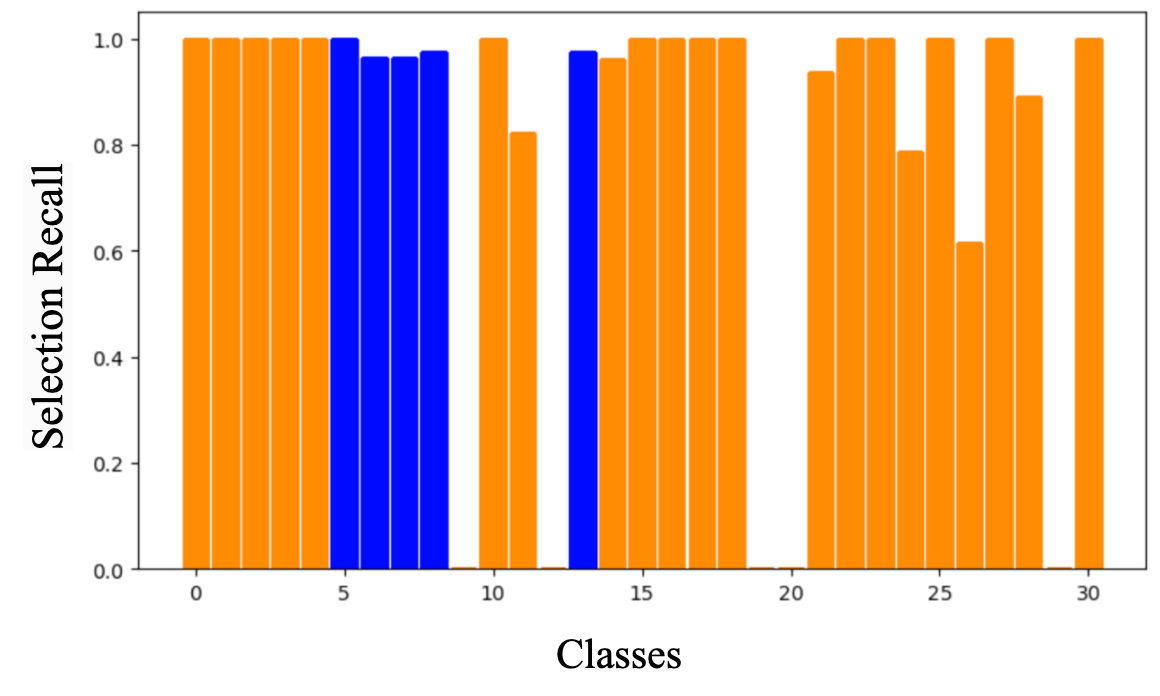}
    \caption{Rec. w/o cls weight}
    \label{fig:exp:recall_wo_cw}
\end{subfigure}
\hfill
\begin{subfigure}[b]{0.2\textwidth}
    \centering
    \includegraphics[width=\textwidth]{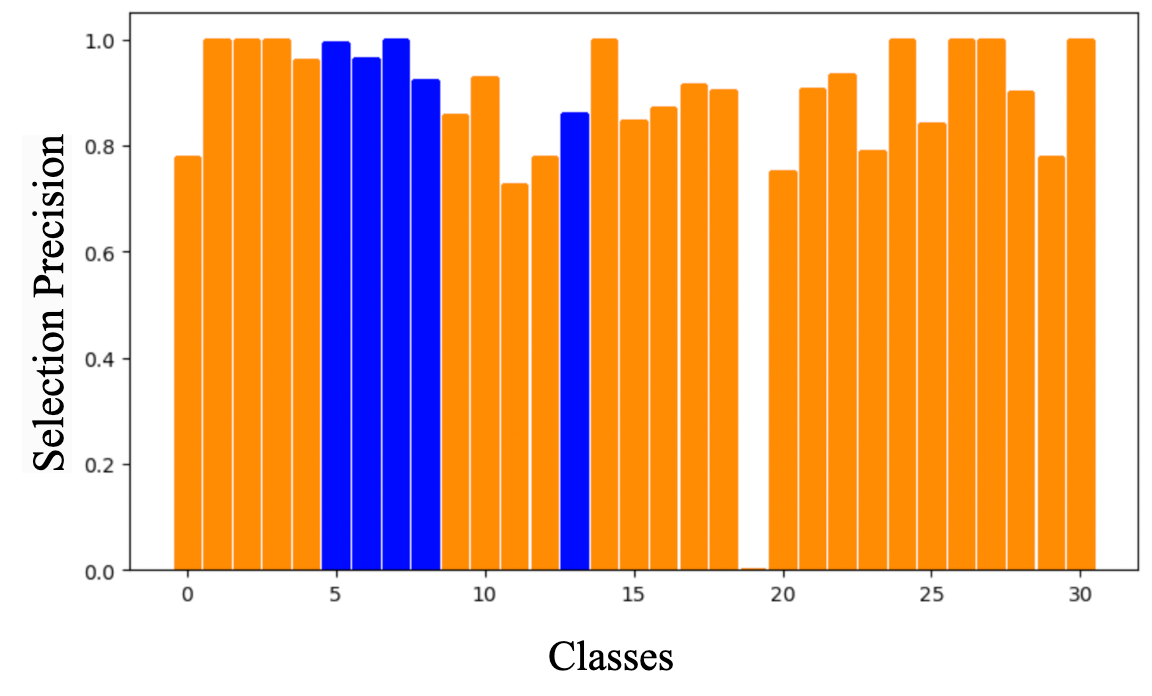}
    \caption{Prec. w/ cls weight}
    \label{fig:exp:precision_w_cw}
\end{subfigure}
\hfill
\begin{subfigure}[b]{0.2\textwidth}
        \centering
        \includegraphics[width=\textwidth]{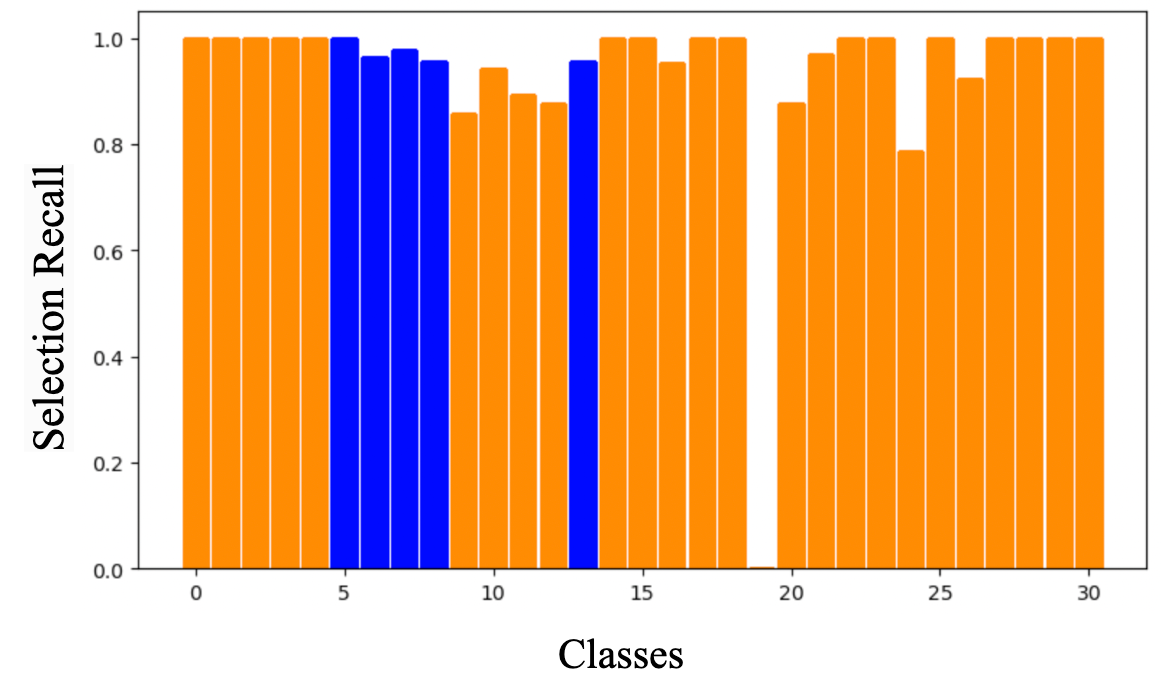}
        \caption{Rec. w/ cls weight}
        \label{fig:exp:recall_w_cw}
    \end{subfigure}
    \caption{Sample selection precision and recall per class on Drive\&Act dataset~\cite{drive_and_act_2019_iccv} after training w/ and w/o class weight. The blue bars represent the head classes, while the orange bars denote the tail classes.}
    \label{fig:exp:class_weight_w/wo}
\end{figure*}
\noindent \textbf{Results.}
Table~\ref{tab:fine_grained} presents the experimental results on the Drive\&Act dataset for fine-grained activity recognition. For comparison, we consider baseline training alongside three representative noisy label learning methods: CDR~\cite{xia2021robust}, ELR~\cite{liu2020earlylearning}, and RRL~\cite{li2021rrl}. These methods, while originally developed for image classification tasks, are adapted to our setting due to the current lack of noisy label learning approaches specifically tailored to driver activity recognition. Across both the MViTv2 and X3D video backbones and under varying noise ratios ($50\%$ and $70\%$), our proposed method consistently achieves superior performance. Notably, the performance gains are particularly prominent when using the lightweight X3D backbone, indicating the effectiveness of our method in settings with limited model capacity.

To further evaluate the generalizability of our approach, we conduct experiments on coarser recognition tasks by processing 90-frame video segments to increase temporal coverage. As reported in Table~\ref{tab:coarse_and_atomic}, our method outperforms all baselines across both the validation and test sets for coarse scenario classification. Specifically, our model surpasses ELR by approximately 3.0\% on the test set, highlighting its capability to generalize to higher-level semantic understanding. For atomic action unit triplet recognition, our approach achieves the best performance in both action and location classification while ranking second for object recognition. The aggregated results on triplet prediction show consistent improvements in overall accuracy, demonstrating the strength of our framework in modeling structured and fine-grained behaviors even under noisy supervision.

The robustness of our method becomes particularly evident under extreme noise conditions. Under the $70\%$ label corruption using the X3D backbone, our model achieves a test accuracy of $53.51\%$, significantly outperforming RRL, which achieves $48.47\%$. This considerable performance gap underscores the resilience of our method against severe annotation noise, a common issue in real-world datasets.

The proposed method enhances performance by learning clustering-friendly spatiotemporal representations through contrastive learning, which facilitates better feature separability under noisy supervision. A novel sample selection strategy, free from manual hyperparameter tuning, allows reliable identification of clean samples while maintaining class balance during training. Additionally, the integration of co-refinement within clusters and soft pseudo-labeling enables effective utilization of noisy data, significantly boosting model robustness and generalization across various recognition tasks. 

These results collectively affirm the effectiveness of our proposed approach in learning noise-robust and semantically meaningful representations. By leveraging spatiotemporal contrastive learning, clustering-based co-refinement, and a hyperparameter-free, class-balanced sample selection strategy, our method achieves strong generalization across various recognition granularities. It successfully mitigates the detrimental effects of noisy labels and provides a scalable solution for real-world driver behavior recognition tasks.

\subsection{Ablation Studies}
\label{exp:ablation_studies}

    Table~\ref{tab:ablation_study} exhibits the ablation studies under $50\%$ label noise to verify the effectiveness of two primary components of our method: Unsupervised Clustering-friendly Representation Learning (UCRL) and class weight. The correction metric measures the proportion of accurate corrections to the total number of noisy labels. These results substantiate that both components are crucial for our proposed method, while class weight plays a more significant role in the sample selection process and model performance compared to UCRL.

    For further qualitative analysis, we visualize the learned representations of training samples either w/ or w/o applying UCRL. The class distributions of head and tail classes are illustrated separately in Figure~\ref{fig:exp:ucrl_w/wo}. 
    Both head classes and tail classes exhibit stronger cohesion within each cluster and clearer boundaries among clusters after training with UCRL. 
    These results demonstrate that UCRL significantly enhances the separability and compactness of the learned representations of both head and tail classes, which is crucial for the subsequent clustering task. 
    
    In Figure~\ref{fig:exp:class_weight_w/wo}, we illustrate sample selection precision and recall per class to verify the effectiveness of class weight.
    We observe that both the precision and recall of several tail classes (containing fewer than $20$ samples) are zero, indicating that none of the clean samples of these tail classes can be filtered without applying class weight (Figure~\ref{fig:exp:precision_wo_cw} \& \ref{fig:exp:recall_wo_cw}). In contrast, we also find that training with class weight leads to more balanced precision and recall across all classes. 
     With the assistance of class weight, the trained model is capable of more accurately selecting clean samples for these tail classes, while maintaining the selection precision and recall for other classes. 
    These results substantiate that class weight alleviates class imbalance issues and improves the sample selection performance, especially for tail classes.

\section{Conclusion}
We propose a robust framework for video-based driver activity recognition under noisy labels, integrating clustering-friendly representation learning, hyperparameter-free sample selection, and adaptive class balancing. Our method outperforms existing approaches on the Drive\&Act benchmark across various recognition tasks. Ablation studies confirm the effectiveness of each component. This efficient, noise-resilient solution supports real-world driver behavior analysis and can be extended to multi-modal or cross-domain applications.

\bibliographystyle{unsrt}
\bibliography{main.bib}

\end{document}